\documentclass[9pt]{article}
\usepackage{spconf,amsmath,amssymb,graphicx,hyperref,booktabs,multirow}
\usepackage{caption}
\usepackage{etoolbox}

\AtBeginEnvironment{thebibliography}{
  \setlength{\itemsep}{0pt}
  \setlength{\parskip}{0pt}
  \setlength{\parsep}{0pt}
}

\usepackage{etoolbox}

\AtBeginEnvironment{thebibliography}{
\setlength{\baselineskip}{11pt}
}
\newcommand{\method}{FD-AA}

\title{\method{}: A Lightweight Focal-Diffuse And Attenuation-Aware Head for Incidental Abdominal Abnormality Detection in Chest CT}
\name{
\begin{tabular}{c}
{\fontsize{11}{10}\selectfont
Haoyan Ding$^{1,2}$, Kritika Iyer$^{1}$,Halid Yerebakan$^{1}$, Zhenyu Bu$^{1}$, Chushu Shen$^{1,2}$, Peiyu Duan$^{1}$, 
Xinyuan Zheng$^{1}$,}\\
{\fontsize{11}{10}\selectfont
Sepehr Farhand$^{1}$, 
Xueqi Guo$^{1}$, 
Chaowei Wu$^{1}$, 
Yoshihisa Shinagawa$^{1}$, 
Gerardo Hermosillo Valadez$^{1}$}
\\
\end{tabular}
\vspace{-4pt}
}

\address{
\begin{tabular}{c}
{\fontsize{11}{9.5}\selectfont
$^{1}$Siemens Medical Solutions USA, Inc., Malvern, PA, USA}\\
{\fontsize{11}{9.5}\selectfont
$^{2}$University of California, Los Angeles, Los Angeles, CA, USA} 
\end{tabular}
}
\begin{document}
\maketitle

\begin{abstract}
Routine chest CT captures upper-abdominal structures that may contain clinically relevant incidental abnormalities. Detecting these findings requires feature extraction from organs with different spatial extents and attenuation patterns. We propose FD-AA, a lightweight organ-aware classification head adaptable for frozen 3-D CT encoders. Within each organ, an attenuation-aware module preserves sparse focal evidence, while masked generalized-mean pooling captures diffuse anomaly patterns. 
In seven abdominal organs, FD-AA with Pillar-0 achieved state-of-the-art (SOTA) performance in both the CT-RATE test set (AUC = 0.798) and the external RAD-ChestCT dataset (AUC = 0.713). More specifically, FD-AA improved macro AUC/AP from 0.763/0.346 to 0.798/0.405 over direct classification using frozen Pillar-0 only (p = 0.034/0.016). Such performance gain generalizes across multiple frozen encoders (AUC improvement on MedicalNet +9.8\%, CT-CLIP +14.7\%, ResNet +3.7\%), demonstrating the effectiveness of FD-AA across different feature representations.
These results support the effectiveness of integrating focal-diffuse aggregation with explicit HU evidence for incidental abdominal abnormality detection.
\end{abstract}


\section{Introduction}
\label{sec:intro}
Routine chest computed tomography (CT) is commonly performed for lung cancer screening and often captures portions of the upper abdomen, where incidental abnormalities may be detected.~\cite{dyer2023incidental,penha2022incidental}. Automatically identifying abdominal abnormalities offers an opportunity to extract additional clinical value without additional scanning or radiation exposure. 

Recent CT foundation models provide a promising starting point, but imaging domain and task discrepancies limit their direct application to this setting. Representative models include CT-CLIP~\cite{hamamci2024ctrate}, developed for chest CT interpretation, and Merlin~\cite{blankemeier2026merlin}, developed using abdominal CT. Both use scan-level global representations for abnormality prediction, but such global aggregation can dilute localized abnormalities within individual abdominal organs. Anatomy-level approaches such as ACA~\cite{kenia2026aca} and fVLM~\cite{shui2025fvlm} offer more localized representations, but do not explicitly separate sparse focal evidence from diffuse organ-wide changes.  
This distinction is particularly relevant to abdominal findings in chest CT, where abnormalities vary substantially in spatial extent and appearance and organs may be only partially captured.~\cite{pavan2022focal,zhang2024tiny} Furthermore, aggregating learned visual features alone does not explicitly preserve quantitative Hounsfield-unit (HU) intensities. To adapt these massive foundation models efficiently, these observations motivate a lightweight organ-aware head that combines complementary spatial pooling with direct HU-derived cues.

Therefore, 
we propose FD-AA to capture organ-level abnormality features according to spatial extent and local intensity patterns.
Our contributions are threefold. First, we combine a sparse focal pathway with a diffuse generalized-mean (GeM) pathway to aggregate localized and distributed organ evidence. Second, we
introduce an attenuation-aware module that explicitly represents both high- and low-attenuation changes within each organ and discounts common extra-organ
confounders to refine the focal pathway. Third, we evaluate on CT-RATE~\cite{hamamci2024ctrate} and perform
external validation on RAD-ChestCT~\cite{draelos2021radchestct}. The proposed head improves
ranking metrics while leaving the 3-D encoder frozen. Using Pillar-0 as the backbone, it established SOTA on CT-RATE and RAD-ChestCT,  outperforming prior supervised baselines and competitive CT-VLMs.

\begin{figure*}[t]
    \centering 
    \includegraphics[width=1\textwidth]{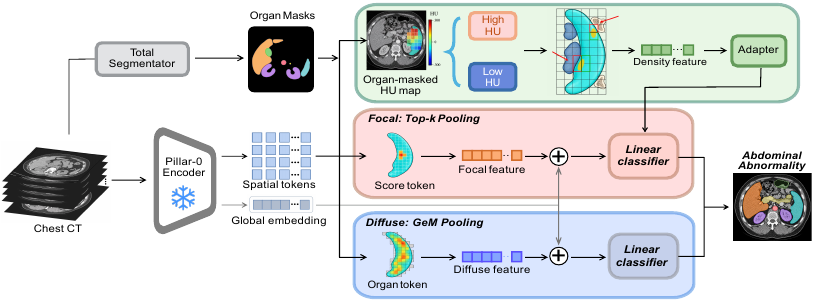}
    \caption{Overview of \method{}. A frozen Pillar-0 encoder extracts spatial
    tokens and a global embedding from the input chest CT, while
    TotalSegmentator supplies organ masks. For each organ, focal pathway uses Top-$k$ pooling to retain sparse focal evidence, whereas diffuse pathway uses GeM to summarize spatially distributed changes. Both representations are
    combined with global context. Besides, organ-masked HU maps encode high- and low-attenuation evidence
    while discounting adjacent intensity confounders. A lightweight adapter
    injects this feature as a residual correction to the
    focal pathway.}
    \label{fig:overview}
\end{figure*}
\section{Method}
\label{sec:method}

\vspace{-2pt}
\subsection{Problem Definition and Preliminaries}
Our goal is to detect abnormalities in upper-abdominal organs incidentally captured in chest CT. Given a chest CT volume $x$, the task is multi-label prediction for seven
abdominal targets $o\in\mathcal O$: abdominal tissue, adrenal gland, aorta,
gallbladder, kidney, liver, and spleen. As shown in
Fig.~\ref{fig:overview}, a frozen 3-D encoder $E$ produces $N=16^3$ spatial
tokens $T=E(x)\in\mathbb{R}^{N\times d}$ and a global embedding $g$.
We first project the two representations as
$h_i=W_t\mathrm{LN}(T_i)\in\mathbb{R}^{d_h}$ and
$\bar g=\mathrm{GELU}(W_g\mathrm{LN}(g))\in\mathbb{R}^{d_h}$, where
$d_h=256$.
Voxel-level
organ masks generated by TotalSegmentator~\cite{wasserthal2023totalsegmentator} are aligned with the input volume and max-pooled onto the
token grid to form $M_o\in\{0,1\}^{N}$. Consequently, a token belongs to an
organ if any voxel in its receptive cell belongs to that organ.

\vspace{-8pt}
\subsection{Focal-Diffuse Organ-Aware Head}

Using the projected features and organ masks defined above, the head first extracts focal features and diffuse features for each organ through two complementary pathways, and each feature group is separately concatenated with the global embedding and processed by a shared MLP to incorporate scan-level context. The linear classifiers then produce two logits. Finally, an organ-specific learnable gate combines the two logits to get the organ-level abnormality.

\noindent\textbf{Organ awareness.}
The head is organ-aware at three levels. First, every organ has its own
learnable query and token support $M_o$. Second, both focal and diffuse pooling
are restricted to that support, preventing large unrelated structures from
dominating the organ representation. Third, the classifier weights and the
focal-diffuse mixing coefficients are organ-specific. 
The mask constrains feature aggregation rather than masking the encoder input, so
each selected token still contains contextual information from its receptive
field. 

\noindent\textbf{Focal evidence.}
The focal pathway is intended for sparse findings such as a small lesion or
calcification. For organ $o$, query $q_o^F$ scores the projected tokens,
\begin{equation}
 s_{o,i}^{F}=\frac{(q_o^F)^\top h_i}{\sqrt{d_h}},\qquad i\in M_o.
\end{equation}
where $d_h=256$ is the projected feature dimension. 
Top-k pooling is applied to the organ tokens to retain
$K_o=\max(1,\lceil0.05|M_o|\rceil)$ tokens with the largest scores.\cite{durand2017wildcat} A softmax
is normalized only over this selected set $S_o$, giving
\begin{equation}
\footnotesize
  \quad S_o=\operatorname{TopK}_{i\in M_o}(s_{o,i}^{F}).
\end{equation}
The proportional $K_o$ adapts the sparse support to organ size while retaining
at least one candidate token.

\noindent\textbf{Diffuse evidence.}
Organ enlargement and diffuse parenchymal changes may present as organ-wide morphological and textural alterations. We therefore use masked generalized-mean pooling~\cite{radenovic2019gem} to capture such evidence.
For channel
$c$,
\begin{equation}
\footnotesize
 z_{o,c}^{D}=\left(\frac{1}{|M_o|}\sum_{i\in M_o}
 [\mathrm{ReLU}(h_{i,c})+\epsilon]^{p}\right)^{1/p}.
\end{equation}
The exponent is bounded to $1<p<6$ for numerical stability and initialized at $p=3$. It is learned
jointly, allowing pooling to move between average-like aggregation and a more
selective maximum without introducing another spatial encoder.

\begin{figure*}[!htb]
    \centering 
    \includegraphics[width=0.95\textwidth]{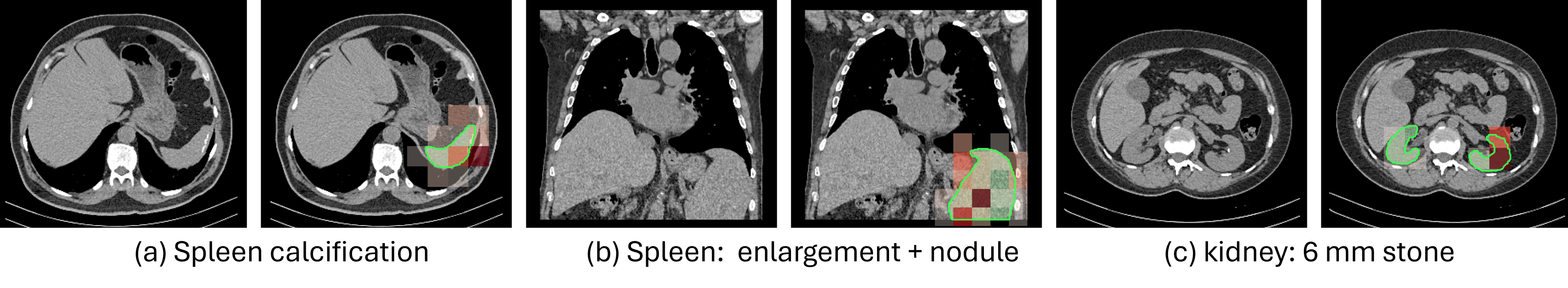}
    \caption{Representative organ-aware attention visualizations. Each case is
    shown as an unannotated CT image (left) and a token-level attention
    overlay with mask (right). Green contours denote the TotalSegmentator organ masks, and the
    colored rectangular patches show token attention, with warmer colors
    indicating greater weight. 
    }
    \label{fig:visualization}
\vspace{-1.5em}
\end{figure*}

\vspace{-8pt}
\subsection{Attenuation-Aware Module}

To compensate for the loss of absolute quantitative intensity cues from the frozen encoder, we compute
token-aligned HU descriptors directly from the input CT.

For high-density evidence, descriptors include within-organ fractions at or above 130, 200, and 400 HU~\cite{agatston1990calcium, han2025density} to capture progressively denser components that may indicate calcifications or calculi.
Besides, to reduce contamination from adjacent bones, within-organ high-HU evidence is contrasted with that in a narrow extra-organ ring.

For low-density evidence, descriptors include fractions at or below 20 and 0 HU, together
with fractions whose attenuation is at least 20, 40, or 80 HU below the
within-organ median~\cite{silverman2019bosniak}. The absolute thresholds provide fluid-attenuation cues, while the relative offsets characterize within-organ attenuation differences. The two directions are combined into a unified evidence map:
\begin{equation}
 e_{o,i}=\max\{e_{o,i}^{\mathrm{high}},e_{o,i}^{\mathrm{low}}\}.
\end{equation}
where \(e_{o,i}^{\mathrm{high}}\) and \(e_{o,i}^{\mathrm{low}}\) denote the confounder-adjusted high- and low-attenuation evidence score for token \(i\) of organ \(o\).

Then this evidence map guides a learned attention to aggregate visual tokens and density evidence, retaining access to surrounding context. A zero-initialized bottleneck adapter projects this feature into the focal feature space as a residual correction, refining the focal prediction without introducing a separate third probability branch. 

\vspace{-8pt}
\subsection{Training Objectives}

The Focal-Diffuse head is trained while keeping $E$ frozen. To address class
imbalance, we use binary cross-entropy loss with organ-wise positive
weights $\omega_o=N_o^-/N_o^+$, where $N_o^+$ and $N_o^-$ are the positive
and negative training counts for organ $o$. We additionally supervise the
global any-abnormality logit:

\begin{equation}
% \footnotesize
 \mathcal L_{FD}=\mathcal L_{\mathrm{WBCE}}
 +\lambda_{\mathrm{any}}\mathcal L_{\mathrm{any}},
\end{equation}
where $\mathcal L_{\mathrm{any}}$ is binary
cross-entropy for the presence of any abdominal abnormality.

The attenuation-aware module is first trained with finding-derived density-change
labels using
{\footnotesize
\begin{align}
 \mathcal L_{H}&=\mathcal L_{\mathrm{sem}}
 +\lambda_{\mathrm{mask}}\mathcal L_{\mathrm{mask}}^H
  +\lambda_{\mathrm{align}}\mathcal L_{\mathrm{align}},\\
 \mathcal L_{\mathrm{mask}}^H&=-\frac{1}{|\mathcal V_H|}
 \sum_{(n,o)\in\mathcal V_H}\log\left(
 \sum_{i:M_{n,o,i}=1}a^H_{n,o,i}\right),
\end{align}
}
where \(\mathcal L_{\mathrm{sem}}\) is weighted binary cross-entropy for predicting whether an organ contains any report-derived high- or low-attenuation finding. The mask loss \(\mathcal L_{\mathrm{mask}}^H\) encourages the attention used for visual feature aggregation to concentrate within the corresponding organ while allowing extra-organ context. Here, \(a^H_{n,o,i}\) is attention normalized over all tokens, and \(\mathcal V_H\) denotes valid scan-organ pairs with active auxiliary supervision.
\(\mathcal L_{\mathrm{align}}\) minimizes KL divergence from the normalized, label-selected evidence map to within-organ attention. When both high- and low-attenuation labels are positive, their evidence maps are combined by an element-wise maximum before normalization. The loss applies only to positive pairs with valid masks and nonempty evidence; labels are not required at inference.
After training the FD head and the attenuation-aware module with their respective objectives, we freeze both and optimize the residual fusion adapter. The final objective combines weighted binary cross-entropy on the fused organ-level logits with an \(L_2\) penalty on the attenuation-induced logit corrections.
Checkpoints are selected by validation AUC. 

\vspace{-2pt}
\section{Experiments and Results}
\label{sec:experiments and results}

\begin{table*}[!htb]
\centering
\small
\setlength{\tabcolsep}{3.2pt}
\caption{Comparison with 3-D CT foundation-model references over seven
abdominal targets. CT-RATE is the internal test cohort ($n=3{,}039$), and
RAD-ChestCT is the external cohort ($n=3{,}630$). Bold
and underlining mark the best and second-best result in each column,
respectively. Groups reflect anatomical modeling.}
\label{tab:sota}
\setlength{\tabcolsep}{5pt}
\begin{tabular}{@{}clcccccccc@{}}
\toprule
& & \multicolumn{4}{c}{CT-RATE (internal test)}
& \multicolumn{4}{c}{RAD-ChestCT (external)}\\
\cmidrule(lr){3-6}\cmidrule(lr){7-10}
 & Method &  AUC$\uparrow$  &  AP  $\uparrow$ & F1$\uparrow$ & Acc.$\uparrow$
& AUC$\uparrow$ & AP $\uparrow$ & F1$\uparrow$ & Acc.$\uparrow$\\
\midrule
\multirow{6}{*}{\rotatebox{90}{Global}} & COLIPRI-CRM~\cite{wald2025colipri}
& \underline{0.761} & \underline{0.357} & 0.415 & 0.579
& \underline{0.644} & \underline{0.227} & 0.285 & 0.310\\
& RadFinder~\cite{ging2026radfinder}
& 0.711 & 0.304 & \underline{0.440} & \underline{0.839}
& 0.612 & 0.218 & \underline{0.316} & \textbf{0.823}\\
& ItemizedCLIP~\cite{lyu2025itemizedclip}
& 0.708 & 0.305 & 0.384 & 0.520
& 0.545 & 0.170 & 0.258 & 0.182\\
& Percival~\cite{beeche2025percival}
& 0.642 & 0.271 & 0.224 & 0.385
& 0.526 & 0.153 & 0.193 & 0.713\\
& Merlin~\cite{blankemeier2026merlin}
& 0.585 & 0.226 & 0.289 & 0.242
& 0.523 & 0.167 & 0.235 & 0.414\\
& CT-CLIP~\cite{hamamci2024ctrate}
& 0.526 & 0.181 & 0.367 & 0.515
& 0.512 & 0.145 & 0.227 & 0.492\\
\midrule
\multirow{6}{*}{\rotatebox{90}{Organ-aware}} & ACA~\cite{kenia2026aca}
& 0.671 & 0.200 &0.292 & 0.649
& 0.530 & 0.157 & 0.239 & 0.537\\
& ViSD-Boost~\cite{cao2025visd}
& 0.571 & 0.212 & 0.307 & 0.236
& 0.521 & 0.160 & 0.241 & 0.178\\
& fVLM~\cite{shui2025fvlm}
& 0.543 & 0.126 & 0.214 & 0.555
& 0.513 & 0.148 & 0.258 & 0.171\\
& Jolia~\cite{khlaut2026jolia}
& 0.522 & 0.187 & 0.256 & 0.409
& 0.505 & 0.145 & 0.265 & 0.475\\
& ARC-CT~\cite{isik2026arcct}
& 0.510 & 0.180 & 0.297 & 0.503
& 0.494 & 0.146 & 0.307 & 0.585\\
& \textbf{\method{} (ours)}
& \textbf{0.798} & \textbf{0.405} & \textbf{0.519} & \textbf{0.878}
& \textbf{0.713} & \textbf{0.336} & \textbf{0.430} & \underline{0.768}\\
\bottomrule
\end{tabular}
\vspace{-1.5em}
\end{table*}

\vspace{-2pt}
\subsection{Datasets and Evaluation}
We evaluate our method on two public datasets.
\textbf{CT-RATE.} This dataset contains non-contrast chest CT scans paired with radiology reports. After preprocessing and quality control, our cohort contains 50,178 volumes: 37,562 for training, 9,577 for validation, and 3,039 for testing, with no patient overlap across splits. Labels are derived from structured radiology reports~\cite{zhang2024radgenome}. \textbf{RAD-ChestCT~\cite{draelos2021radchestct}.} For external validation, we use all 3,630 scans from the public release. Its abnormality--location matrix is mapped to the same seven abdominal targets. Due to class imbalance, we report macro AUC and AP as primary metrics, together with F1, precision, recall, label-wise accuracy, and Any-AUC. Any-AUC measures scan-level detection of any abdominal abnormality.
\vspace{-0.8cm}
\subsection{Implementation Details}
The default encoder is the frozen abdominal CT Pillar-0~\cite{agrawal2025pillar0}. Its precomputed $16^3$ spatial tokens are reused across head experiments. Token, global, and branch features are projected to 256 dimensions, and the attenuation-aware module uses a 64-dimensional bottleneck adapter. The focal pathway retains the top 5\% of organ tokens, and GeM pooling is initialized with $p=3$. We set \(\lambda_{\mathrm{any}}=0.2\), \(\lambda_{\mathrm{mask}}=0.05\), and \(\lambda_{\mathrm{align}}=0.1\). The Focal-Diffuse head and attenuation-aware module are optimized with AdamW using a learning rate of $3\times10^{-4}$ and weight decay of $10^{-4}$. The residual-fusion adapter uses a learning rate of $5\times10^{-4}$ and weight decay of $10^{-3}$.

\vspace{-8pt}
\subsection{Evaluation Across Image Encoders}
We evaluate the proposed classification head with four frozen encoders: Pillar-0, MedicalNet~\cite{chen2019med3d}, ResNet-18~\cite{he2016deep}, and CT-CLIP~\cite{hamamci2024ctrate}. Table~\ref{tab:encoders} compares four ablative configurations: \textbf{GA (global-aware)}, which directly predicts from a scan-level global embedding of the encoder; \textbf{OA (organ-aware)}, which extracts organ-specific features using queries and mask-guided attention; \textbf{FD (focal-diffuse)}, which combines sparse top-\(k\) focal pooling with organ-wide GeM pooling; and \textbf{FD-AA (ours)}, which further incorporates attenuation-aware features. FD-AA consistently improves AUC and AP across all evaluated encoders, showing that the proposed head generalizes across different representations. Pillar-0 achieves the strongest overall performance. For each encoder, the ablation study in the four configurations further validates the component-wise contributions of organ-aware feature extraction, focal-diffuse aggregation, and attenuation-aware modeling. Fig.~\ref{fig:visualization} shows qualitative examples of localized attenuation-related responses and broader organ-level responses to visually apparent abnormalities. Overall, focal pooling preserves sparse evidence, diffuse pooling captures organ-wide abnormalities, and attenuation-aware modeling complements learned features with explicit intensity cues.

\begin{table}[!htb]
\centering
\scriptsize
\setlength{\tabcolsep}{0.845pt}
\caption{Evaluation Across Image Encoders on CT-RATE. Rows within
each encoder compare classification heads. 
}
\label{tab:encoders}
\resizebox{\columnwidth}{!}{
\begin{tabular}{@{}llccccccc@{}}
\toprule
Encoder & Method & AUC & AP & F1 & Prec. & Rec. & Acc. & Any-AUC\\
\midrule
Pillar-0 & GA & 0.763 & 0.346 & 0.428 & 0.348 & 0.557 & 0.835 & 0.805\\
 & OA & 0.768 & 0.355 & 0.461 & 0.389 & 0.568 & 0.853 & 0.809\\
 & FD & 0.790 & 0.395 & 0.514 & 0.449 & \textbf{0.602} & 0.874 & 0.826\\
 & FD-AA & \textbf{0.798} & \textbf{0.405} & \textbf{0.519} & \textbf{0.461} & 0.594 & \textbf{0.878} & \textbf{0.838}\\
\midrule
MedicalNet & GA & 0.682 & 0.213 & 0.296 & 0.198 & 0.587 & 0.690 & 0.733\\
& OA & 0.695 & 0.250 & 0.350 & 0.254 & 0.572 & \textbf{0.765} & 0.762\\
 & FD & 0.710 & 0.260 & 0.323 & 0.211 & 0.685 & 0.681 & 0.750\\
 & FD-AA& \textbf{0.749} & \textbf{0.329} & \textbf{0.375} & \textbf{0.256} & \textbf{0.695} & 0.743 & \textbf{0.816}\\
 \midrule
CT-CLIP & GA & 0.645 & 0.185 & 0.279 & 0.176 & 0.672 & 0.616 & 0.704\\
& OA & 0.673 & 0.196 & 0.327 & 0.224 & 0.608 & 0.723 & 0.732\\
 & FD & 0.668 & 0.214 & 0.295 & 0.187 & \textbf{0.706} & 0.626 & 0.752\\
 & FD-AA & \textbf{0.740} & \textbf{0.319} & \textbf{0.377} & \textbf{0.260} & 0.684 & \textbf{0.749} & \textbf{0.817}\\
\midrule
ResNet-18 & GA & 0.705 & 0.227 & 0.316 & 0.208 & 0.655 & 0.686 & 0.763\\
& OA & 0.701 & 0.232 & 0.373 & \textbf{0.282} & 0.552 & \textbf{0.795} & 0.750\\
 & FD & 0.709 & 0.242 & 0.294 & 0.183 & \textbf{0.741} & 0.604 & 0.767\\
 & FD-AA & \textbf{0.731} & \textbf{0.279} & \textbf{0.342} & 0.227 & 0.694 & 0.734 & \textbf{0.789}\\
\bottomrule
\end{tabular}
}
\vspace{-1.8em}
\end{table}

\vspace{-8pt}
\subsection{Comparison with State-of-the-Art Methods}
Table~\ref{tab:sota} compares our method with existing 3-D CT foundation-model approaches on the same seven abdominal targets. All results for our method are obtained using the FD-AA head with a frozen Pillar-0 encoder. FD-AA achieves the highest macro AUC, AP, and F1 score on both the internal CT-RATE test set and the external RAD-ChestCT cohort. On CT-RATE, it also achieves the highest label-wise accuracy. The improvements observed on both the internal and external cohorts suggest that the proposed approach generalizes across these two datasets despite the distribution shift.

\vspace{-8pt}
\section{Discussion and Conclusion}
\label{sec:discussion}

This study has some limitations for future work. Report-derived labels may be noisy or omit subtle findings, and the lack of lesion-level annotations limits direct validation of attention localization. FD-AA also relies on organ masks at inference, and abdominal organs may be only partially captured in chest CT. Future work will explore mask-free inference and partial organ coverage.

In conclusion, we presented a lightweight head for incidental abdominal abnormality detection
in chest CT. Separating sparse focal from diffuse organ evidence and adding an
attenuation-aware module improves internal and external ranking metrics with
a frozen 3-D foundation encoder, preserving spatial and quantitative CT evidence without training a separate encoder.

\bibliographystyle{IEEEbib}
\bibliography{refs}
\end{document}